\documentclass[sn-mathphys-num]{sn-jnl}
\usepackage{graphicx}%
\usepackage{multirow}%
\usepackage{amsmath,amssymb,amsfonts}%
\usepackage{amsthm}%
\usepackage{mathrsfs}%
\usepackage[title]{appendix}%
\usepackage{xcolor}%
\usepackage{textcomp}%
\usepackage{manyfoot}%
\usepackage{nomencl}
\usepackage{booktabs}%
\usepackage{rotating}%
\usepackage{listings}%
\usepackage[utf8]{inputenc}
\usepackage[T1]{fontenc}
\usepackage{tabularray}
\usepackage{colortbl}

\usepackage{footnote}
\usepackage{tabularx}      
\usepackage{array}         
\usepackage{adjustbox}
\usepackage{amssymb}       

\theoremstyle{thmstyleone}%
\theoremstyle{thmstyletwo}%

\theoremstyle{thmstylethree}%

\begin{document}

\title[]{Extended KAFR: A kinematic-adaptive paradigm for the efficient analysis of surgical video}


\author*[1]{\fnm{Huu~Phong} \sur{Nguyen}}\email{huuphong.nguyen@utsouthwestern.edu}

\author[1]{\fnm{Shekhar~Madhav} \sur{Khairnar}}\email{
shekharmadhav.khairnar@utsouthwestern.edu}

\author*[1]{\fnm{Ganesh} \sur{Sankaranarayanan}}\email{ganesh.sankaranarayanan@utsouthwestern.edu}

\affil[1]{\orgdiv{Department of Surgery}, \orgname{University of Texas Southwestern Medical Center}, \orgaddress{\street{5323 Harry Hines Blvd}, \city{Dallas}, \postcode{75390}, \state{Texas}, \country{USA}}}


\abstract{
Artificial Intelligence is increasingly applied to surgical video analysis for
phase segmentation, skill assessment, and workflow optimization.
A key challenge is the length of surgical recordings, often one to several hours,
creating substantial computational burden.
We previously developed Kinematics-Adaptive Frame Recognition (KAFR) for robotic
surgery, showing that tracking tool motion effectively identifies informative frames
while filtering redundant content.
However, laparoscopic surgery introduces additional challenges: manual camera control
causes frequent motion artifacts, and image quality is generally lower than robotic
systems.
This study evaluates whether KAFR generalizes to laparoscopic surgery using the
Cholec80 benchmark, comprising 80 laparoscopic cholecystectomy procedures annotated
for seven surgical phases.
KAFR operates in three stages: a fine-tuned YOLO model detects and segments surgical
tools; frames are adaptively selected based on tool displacement or velocity variation;
and an X3D model classifies selected frames into surgical phases.
KAFR achieved a 91.0\% F1 score using only 0.58\% of frames for phase classification,
representing an approximately seven-fold reduction compared to typical 4\% frame
sampling, while maintaining performance comparable to LoViT (90.2\%) and
Trans-SVNet (89.7\%).
These results demonstrate that kinematics-based frame selection transfers effectively
to the challenging laparoscopic environment.
}




\flushbottom
\maketitle
%
%
\thispagestyle{empty}

\section{Introduction}
\label{sec:introduction}
Surgery is essential to modern healthcare, with an estimated 313 million procedures conducted globally each year~\cite{meara2015global}. Surgical outcomes rely on surgeon training, intraoperative decision-making, and continuous performance improvement~\cite{nguyen2026ai,khairnar2025automated,khairnar2025machine,pydimarry2024evaluating,cizmic2025artificial,bain2024artificial,MAIERHEIN2022surgical,garrow2021machine}.  Traditional approaches to surgical education, however, are mostly based on apprenticeship models in which performance assessment remains largely subjective, and real-time coaching during complex procedures is limited to the experience of the operating team. Video recording of surgical procedures has become routine in many institutions, resulting in extensive archives of operative footage with potential value for training and quality assurance~\cite{khanna2025enhancing,li2025surgical,li2024deep,harari2024deep, morris2023deep,lavanchy2023preserving,mascagni2022computer,kitaguchi2022artificial,liu2025deep,
guo2023current}.

Deep learning has led to substantial advances in automated surgical video analysis. Convolutional neural networks (CNN)~\cite{phong2023pattern,nguyen2019advanced,phong2019improvement,phong2017offline,lecun2002gradient} first showed the feasibility of recognizing general video classification~\cite{phong2018action,karpathy2014large}, which were later adapted for surgical phase segmentation~\cite{twinanda2016endonet}. Three-dimensional CNN extended these capabilities to capture motion patterns over time~\cite{tran2015learning}, while recurrent neural networks architectures enabled modeling of temporal dependencies~\cite{carreira2017quo}.Recently, Transformer-based techniques have set new benchmarks by employing multi-head attention to identify complex relationships between distant surgical activities~\cite{liu2025lovit,nguyen2023video,czempiel2021opera}. These methods have advanced the field toward clinically useful applications. However, a practical challenge remains: the computational cost of processing hour-long surgical videos limits deployment in clinical settings where resources are constrained and rapid analysis is critical.

We tackled this challenge by analyzing how expert surgeons review operative footage. Surgeons prioritize decisive measures, such as instrument engagement, crucial structural exposure, and key movements, over other aspects of the process. Instrument repositioning, waiting, or minimal activity receive less attention. This selective attention is both intuitive and efficient. We developed Kinematics-Adaptive Frame Recognition (KAFR) to computationally capture this pattern, selecting frames for analysis based on instrument motion rather than elapsed time~\cite{nguyen2025kinematic}. By tracking surgical instruments and identifying moments of meaningful kinematic activity, KAFR achieves competitive phase segmentation accuracy while analyzing only a small fraction of available frames.

Prior work validated KAFR on robotic surgery, where high-resolution imaging and stable camera positioning facilitate reliable tool tracking~\cite{nguyen2025kinematic}. Laparoscopic surgery presents a more challenging environment: the camera is manually controlled at the bedside, introducing frequent motion artifacts; image quality is generally lower; and instrument appearance varies more substantially across procedures and institutions. Whether motion-based frame selection remains effective under these less controlled conditions remains an open question with important implications for the broader applicability of this approach.

Here we extend KAFR to laparoscopic cholecystectomy and evaluate its performance on the Cholec80 benchmark dataset~\cite{twinanda2016endonet}. Laparoscopic cholecystectomy is among the most commonly performed abdominal procedures, with over 700{,}000 cases yearly in the United States~\cite{abdallah2025difficult}, making it a clinically relevant testbed for validating new analytical methods. 

The remainder of this article is organized as follows: Section~\ref{sec:results} presents the experimental setup, dataset, evaluation metrics, and results. Section~\ref{sec:discussion} discusses implications and limitations. Section~\ref{sec:methods} describes the KAFR framework, including instrument detection, adaptive frame selection strategies, and phase segmentation.

\section{Results}
\label{sec:results}
\subsection{Benchmark Datasets}
\label{sec:dataset}
While earlier evaluations of KAFR were conducted on two robotic surgical datasets including roughly 150 videos~\cite{nguyen2025kinematic}, this study focuses on laparoscopic surgery using Cholec80. Unlike robotic datasets, Cholec80 poses unique challenges, including handheld camera motion, non-rigid organ deformation, and more frequent tool occlusions. This evaluation helps establish the framework's effectiveness under varied visual and kinematic settings.

\begin{figure} [bt!]
\centering
\includegraphics[keepaspectratio,width=0.95\textwidth]{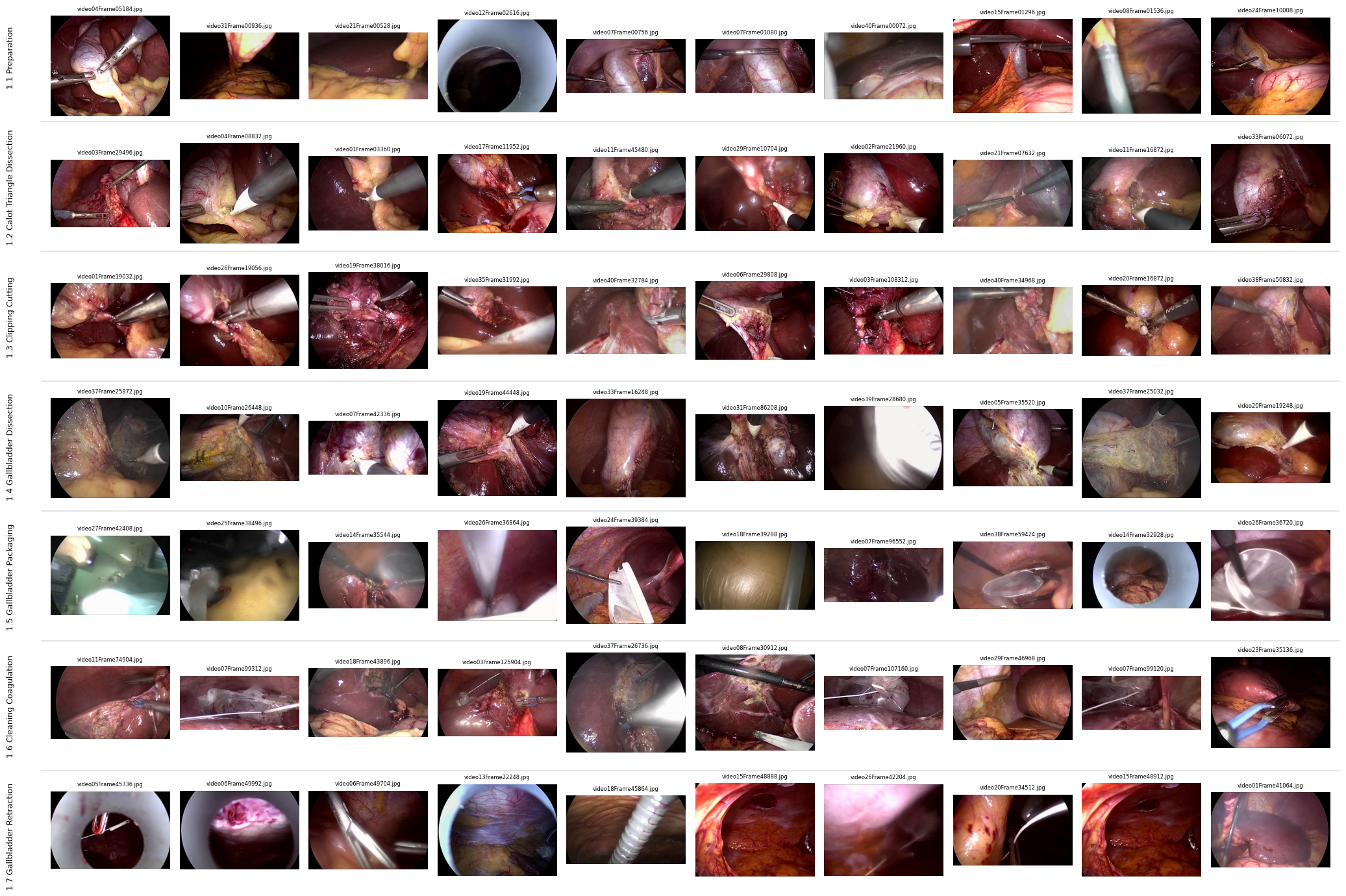}
\caption{Random samples from the Cholec80 dataset illustrating the seven surgical phase categories.}
\label{fig:cholec80_phases}
\end{figure}

The Cholec80 dataset~\cite{twinanda2016endonet} is a publicly available collection of 80 videos capturing real-world laparoscopic cholecystectomy procedures performed by 13 distinct surgeons at the University Hospital of Strasbourg. Each video is recorded at 25 frames per second (FPS) and manually annotated at 1 fps for surgical phase segmentation. The dataset defines seven distinct surgical phases: Preparation (P0), Calot triangle dissection (P1), Clipping and cutting (P2), Gallbladder dissection (P3), Gallbladder packaging (P4), Cleaning and coagulation (P5), and Gallbladder retraction (P6). Annotations were curated by an expert surgeon to ensure clinical accuracy. Cholec80 includes a wide range of inter-procedural variability, making it a robust benchmark for evaluating surgical workflow analysis methods, particularly in temporal modeling and surgical phase segmentation (Figure~\ref{fig:cholec80_phases}).
To align with previous research practices~\cite{nguyen2025kinematic} and reduce computational load due to the considerable length of the videos, the frame rate was first converted to 24 fps and frames were sampled at 6 fps. Of the 80 videos, 40 were used for training, 8 for validation, and 32 for testing. All frames were annotated, resulting in 31{,}125 images for training, 21{,}389 for validation, and 76{,}570 for testing.
\subsection{Object Detection}

\begin{figure} [tbh!]
\centering
\includegraphics[keepaspectratio,width=0.55\textwidth]{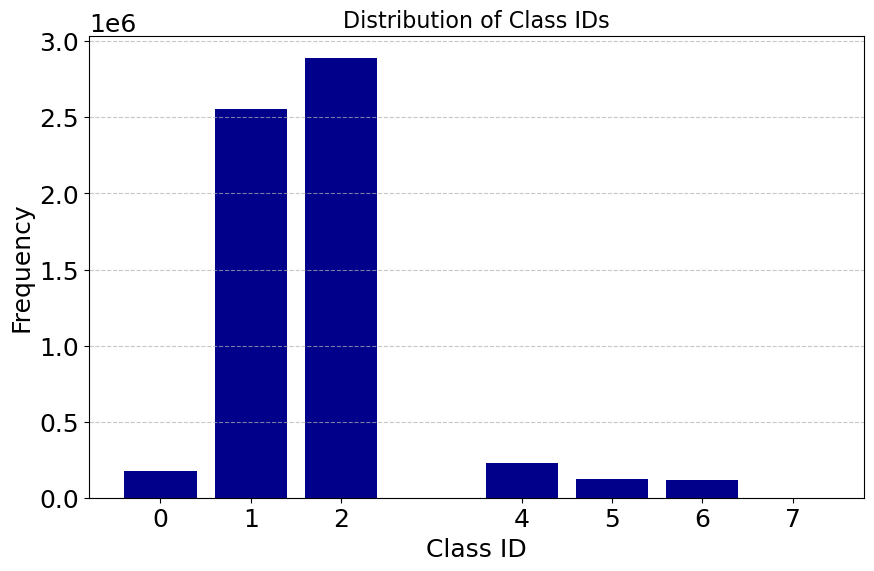}
\caption{Distribution of class IDs for surgical tools. The tools are tracked by the YOLOv8 model, each assigned a unique class identifier.}
\label{fig:distribution}
\end{figure}

Figure~\ref{fig:distribution} illustrates the distribution of frame counts for each class ID, corresponding to the eight surgical instruments listed in Table~\ref{tab:table_toolname} from the Cholec80 dataset. Instrument detection and tracking were performed using YOLOv8, as detailed in Section~\ref{sec:objectdetection}. Notably, "Grasper 2" and "Grasper 3" are functionally identical to "Grasper 1," as they all represent the same type of instrument. Therefore, class IDs 3 and 7 were merged into class ID 2 to reflect this redundancy.

Since some tools are identical to others, to avoid confusion in our analysis of tools for object detection, we define the following naming conventions:
\begin{itemize}
    \item ``One Tool'': Tracking a single Grasper positioned on the right side of the screen.
    \item ``Two Tools'': Tracking two Graspers, one on the left and one on the right side of the screen.
    \item ``Four Tools'': Tracking both a Grasper and a Hook on each side of the screen--one Grasper and one Hook on the left, and one Grasper and one Hook on the right.
\end{itemize}

In addition, the image is divided into two halves by a vertical line at the center. At any given moment in the video, a tool is considered the left-hand tool if its centroid is located on the left half and the right-hand tool if its centroid is located on the right half.
%

\begin{table}[tb!]
\caption{List of surgical tools used in the dataset. Each tool is listed with its unique identifier and functional category. Tools with similar manipulation functions (e.g., graspers) are grouped under the same category. Note: class ID 3 (Grasper 2) and class ID 7 (Grasper 3) were merged into class ID 2 (Grasper 1) due to functional equivalence; the merged class is treated as a single Grasper category in all analyses.}

\label{tab:table_toolname}
\begin{tabular}{lll}
\hline
\textbf{Tool ID} & \textbf{Tool Name} & \textbf{Functional Category} \\
\hline
0 & Bipolar    & Bipolar   \\
1 & Hook       & Hook      \\
2 & Grasper 1  & Grasper   \\
3 & Grasper 2  & Grasper   \\
4 & Irrigator  & Irrigator \\
5 & Clipper    & Clipper   \\
6 & Scissors   & Scissors  \\
7 & Grasper 3  & Grasper   \\
\hline
\end{tabular}
\end{table}

To evaluate the reliability of tool detection, we examined the fine-tuned YOLOv8 model on the Cholec80 dataset.
Annotation frames were drawn exclusively from the Cholec80 training split; frames with visible blur were excluded, and the validation and test splits used for phase segmentation were withheld entirely to prevent data leakage. Frames were sampled from across the training videos to ensure coverage of all seven surgical phases and all instrument types, with candidate frames selected to maximize diversity across different procedural stages and videos.
The resulting 750 annotated frames, labeled with segmentation masks, were divided into 584 training images, 78 validation images, and 88 test images using a stratified 80/10/10 split.
Grasper variants originally labeled as Grasper 1, 2, and 3 were merged into a single Grasper class, yielding six instrument classes in total.
\begin{table}[tb!]
\caption{Per-class annotation instance counts for the 750-frame annotated subset used to fine-tune the YOLOv8 tool detector. Because a frame can contain multiple tools or multiple instances of the same tool, these values refer to annotation instances rather than the number of frames containing each tool. Grasper variants (originally Grasper 1, 2, and 3) were merged into a single Grasper class, yielding six instrument classes in total.}
\label{tab:annotation_counts}
\begin{tabular}{lcccc}
\hline
\textbf{Class Name} & \textbf{Train} & \textbf{Val} & \textbf{Test} & \textbf{Total} \\
\hline
Grasper   & 430 & 59  & 57  & 546  \\
Hook      & 227 & 49  & 65  & 341  \\
Bipolar   & 85  & 6   & 4   & 95   \\
Irrigator & 49  & 3   & 2   & 54   \\
Scissors  & 59  & 6   & 6   & 71   \\
Clippers  & 107 & 9   & 7   & 123  \\
\hline
\textbf{Total} & \textbf{957} & \textbf{132} & \textbf{141} & \textbf{1,230} \\
\hline
\end{tabular}
\end{table}

The detector achieved overall box precision of $0.894$, recall of $0.912$, mAP@50 of $0.934$, and mAP@50:95 of $0.877$.
Frequently used instruments such as Hook and Bipolar achieved very high precision and recall, both greater than $0.98$.
Rarer tools such as Clipper ($0.831$ and $0.714$) and Scissors ($0.735$ and $1$) showed comparatively lower but still reliable performance. Mask-based evaluation further demonstrated robustness with an overall mAP@50:95 of $0.827$.
Expanding the annotated dataset in future work would further strengthen the evaluation of the detection component.
\subsection{Hyperparameter}
Our proposed methods were developed using a \texttt{PyTorch} framework and implemented on a Windows\textsuperscript{\textregistered} 11 Enterprise system equipped with an AMD Ryzen\textsuperscript{TM} Threadripper\textsuperscript{TM} PRO 3995WX 64-cores CPU @ 2.70 GHz. We trained our model with two Nvidia RTX A6000 GPUs.  

The experiments were performed based on the configurations specified in Table~\ref{tab:table_hyperparameter}. The hyperparameters such as image size, batch size, and clip length, were optimized to maximize GPU memory usage. The threshold values for KAFR are determined based on the desired number of extracted frames, such that instead of selecting a value between 0 and 1 as in the Equation~\ref{eq:f1}, we choose a number of frames by a percentage of the total number of training samples and adjust the threshold adaptively. The beta exponents $\beta_d$ and $\beta_v$ were both set to 1 in this study as a pragmatic choice. A joint search over beta and the other hyperparameters would require an exponentially growing number of experimental configurations, making it impractical within the scope of this current study. A more extensive strategy of hyperparameter optimization will be explored in future work. Experiments were repeated three times, and the arithmetic mean and standard deviation of performance metrics were computed. Throughout training, model checkpoints were monitored based on the best validation F1 score, and the final reported results correspond to the model achieving the highest F1 score on the testing set.

\begin{table}[tb!]
\centering
\caption{Hyper-parameter Settings and Descriptions. Hyper-parameter settings used consistently throughout the article. Any deviations from these settings are explicitly indicated in the corresponding sections.}
\label{tab:table_hyperparameter}
\begin{tabular}{|l|l|p{3.5cm}|p{6.5cm}|}
\hline
Category & Method & Value & Description \\
\hline
\multirow{10}{*}{Model Parameter} 
& Image size & $300 \times 300$ & Input image dimensions \\
& Optimizer & AdamW & Optimization algorithm \\
& Schedule & StepLR & Learning rate scheduler \\
& Learning rate & 0.001 & Initial learning rate \\
& Decay rate & 0.7 & Decay applied to learning rate \\
& Batch size & 64 & Number of samples per batch \\
& Epochs & 100 & Total training epochs \\
& Early stop & 10 & Early stopping patience \\
& Clip length & 16 & Number of frames in each clip \\
& Median & 114 & Median number of frames per phase segment used during resampling; computed separately from batch assembly \\
& X3D backbone & Kinetics-400 & Fine-tuned from pre-trained Kinetics-400 checkpoint \\
\hline
KAFR & Threshold & Flexible & Motion threshold (adaptive) \\
\hline
\multirow{5}{*}{Augmentation} 
& RandomResizedCrop & scale=(0.8, 1.0), size=224 & Zoom in and out by cropping \\
& ColorJitter & brightness=0.2, contrast=0.2, saturation=0.2, hue=0.1 & Random brightness, contrast, saturation, hue adjustment \\
& GaussianBlur & kernel\_size=3 & Apply Gaussian blur \\
& RandomErasing & p=0.5 & Randomly erase portions of the image \\
& RandomPerspective & distortion\_scale=0.5, p=0.5 & Apply random perspective transformation \\
\hline
\multirow{2}{*}{Preprocessing} & Butterworth Filter & cutoff=3Hz, fs=24Hz & Apply low-pass Butterworth filter to x and y coordinates \\
& SMA window & 2--200 (post-hoc) & Smooth Moving Average window size; evaluated as a post-hoc sensitivity analysis on test-set predictions \\
\hline
\end{tabular}
\end{table}

While we tuned major hyper-parameters for optimal performance, a full exhaustive search was not conducted due to the exponentially growing design space as the number of parameters increases. As a result, our hyperparameter tuning was guided by prior studies, empirical heuristics, and the constraints imposed by available hardware resources.

In these experiments, we evaluated two adaptive variants of the KAFR approaches, namely Adaptive 1 and Adaptive 2, each incorporating a different temporal representation of tool motion to assess the sensitivity of our framework to dynamic changes. Adaptive 1 computes cumulative tool velocity across a temporal window to determine motion intensity, prioritizing frames with the highest aggregate movement. Adaptive 2 builds on this concept by using cumulative acceleration instead of velocity, thereby focusing on abrupt changes in surgical tool dynamics.

Since tool kinematics derived from image-based detection are sensitive to visual variability, particularly when estimating velocity and acceleration, we evaluated our approaches under two experimental settings aimed at enhancing robustness. To accommodate the highly dynamic and visually complex nature of laparoscopic surgery, additional mechanisms were incorporated to mitigate the impact of noise and instability that could arise from frame reduction. \emph{Setting 1} focuses on visual augmentation, applying frame-level transformation to improve resilience against camera motion, lighting changes, and occlusion. \emph{Setting 2} builds on this by adding temporal filtering to refine kinematic signal and stabilize motion patterns.
\subsection{Evaluation Metrics}
We evaluate phase classification performance using standard metrics: accuracy, precision, recall, and F1 score, computed across all surgical phases. For multi-class evaluation, we report macro-averaged precision, recall, and F1 score to account for class imbalance across phases.

In addition to strict frame-level evaluation, we apply the 10-second relaxed boundary protocol~\cite{liu2025lovit}. This approach tolerates small temporal misalignments near phase transitions, where predictions within a 10-second window at segment boundaries are permitted to differ by one or two phase indices. This relaxation reflects the inherent ambiguity in defining exact phase transition points during surgery.

Relative performance changes between methods are reported as percentage change: $\Delta = (\text{New} - \text{Original}) / \text{Original} \times 100\%$.

\subsection{Comparison with state-of-the-art methods}
\label{sec:effectiveness}

To validate the effectiveness and robustness of our proposed method, we conducted a comprehensive comparison between KAFR and several state-of-the-art approaches, including LoViT, MS-AST, and Trans-SVNet, as summarized in Table~\ref{tab:comparison}. These methods represent recent advancements in surgical phase segmentation and have demonstrated strong performance on the Cholec80 benchmark dataset.

Benchmarked against EndoNet, the pioneer deep learning model applied to laparoscopic video analysis, KAFR demonstrates substantial improvements. EndoNet uses a 2-D CNN architecture that processes frames independently, limiting its ability to capture temporal dependencies essential for phase segmentation. On the Cholec80 benchmark, EndoNet reports an F1 score of 76.5\% and accuracy of 81.7\% under the 10-second relax setting. In contrast, KAFR achieves 91.0\% F1 and 91.5\% accuracy, representing improvements of 19.0\% and 12.0\%, respectively. Crucially, this performance gain is coupled with a massive reduction in computational overhead: KAFR selects only 0.58\% of video frames for phase classification, compared to the 4\% used by EndoNet --- a 7-fold reduction in frames processed by the classification module.

Compared to Trans-SVNet, a Transformer-based architecture, KAFR achieves competitive F1 score (91.0\% vs. 89.7\%) and accuracy (91.5\% vs. 90.3\%), with comparable recall and precision. This result suggests that motion-based frame selection can be competitive with more complex Transformer architectures.

Although MS-AST reports an accuracy of 94.5\%, KAFR achieves competitive F1 score (91.0\% vs. 90.1\%), precision (90.5\% vs. 90.4\%), and recall (91.6\% vs. 90.0\%). The competitive F1 score suggests balanced performance across phases.

In relation to LoViT, the state-of-the-art model on F1 score, KAFR achieves competitive F1 score (91.0\% vs. 90.2\%) and comparable precision (90.5\% vs. 89.9\%) and recall (91.6\% vs. 90.6\%), despite a slightly lower accuracy (91.5\% vs. 92.4\%). These results suggest balanced prediction performance across phases.

For comparison under strict evaluation (without 10-second relaxation), we used a baseline KAFR model with 1.5\% frame retention. This model achieves 85.1\% F1 score and 88.9\% accuracy, using only one-third of the frames required by existing methods, which remains competitive with LoViT and Trans-SVNet. These results demonstrate KAFR's robustness across evaluation protocols.
\begin{table}[hb!]
\centering
\caption{Performance (\%) comparison of state-of-the-art methods on the Cholec80 dataset. F1 scores (for other methods if unavailable) are computed from the corresponding precision and recall values. The results for Trans-SVNet and AVT were provided by~\cite{liu2025lovit} under the non-relaxed setting. Other methods apply uniform downsampling (typically to 1 fps, i.e., 4\% of original frames). The KAFR baseline incorporates uniform downsampling and data balancing, while the KAFR method adds adaptive frame selection based on tool motion. The baseline model uses all frames in Setting 1. The best methods are highlighted in bold.}
\label{tab:comparison}
\begin{tabular}{lcccccc}
\toprule
\textbf{Method} & \textbf{10-sec relax} & \textbf{Accuracy} & \textbf{Precision} & \textbf{Recall} & \textbf{F1 score} & \textbf{Retained Frames (\%)} \\
\midrule
EndoNet (ITMI'16)~\cite{twinanda2016endonet}             & \checkmark & 81.7\,$\pm$\,4.2 & 73.7\,$\pm$\,16.1 & 79.6\,$\pm$\,7.9 & 76.5\,$\pm$\,16.5 & 4 \\
SV-RCNet (ITMI'17)~\cite{jin2017sv}                      & \checkmark & 85.3\,$\pm$\,7.3 & 80.7\,$\pm$\,7.0 & 83.5\,$\pm$\,7.5 & 82.0\,$\pm$\,6.9 & 20 \\
TeCNO (MICCAI'20)~\cite{czempiel2020tecno}               & \checkmark & 88.5\,$\pm$\,0.2 & 81.6\,$\pm$\,0.4 & 85.2\,$\pm$\,1.0 & 83.3\,$\pm$\,0.3 & 20 \\
MTRCNet-CL (MIA'20)~\cite{jin2020multi}                  & \checkmark & 89.2\,$\pm$\,7.6 & 86.9\,$\pm$\,4.3 & 88.0\,$\pm$\,6.9 & 87.4\,$\pm$\,4.3 & 4 \\
UATD (ITMI'20)~\cite{ding2023less}                       & \checkmark & 91.9\,$\pm$\,5.6 & 89.5\,$\pm$\,4.4 & 90.5\,$\pm$\,5.9 & 89.9\,$\pm$\,4.4 & 4 \\
Trans-SVNet (MICCAI'21)~\cite{gao2021trans}              & \checkmark & 90.3\,$\pm$\,7.1 & 90.7\,$\pm$\,5.0 & 88.8\,$\pm$\,7.4 & 89.7\,$\pm$\,5.0 & 4 \\
Swin-BiGRU (PMLR'22)~\cite{he2022empirical}              & \checkmark & 93.8\,$\pm$\,0.0 & 89.9\,$\pm$\,0.7 & 89.6\,$\pm$\,0.5 & 89.7\,$\pm$\,0.6 & -- \\
TMRNet (ITMI'21)~\cite{jin2021temporal}                  & \checkmark & 90.1\,$\pm$\,7.6 & 90.3\,$\pm$\,3.3 & 89.5\,$\pm$\,5.0 & 89.8\,$\pm$\,3.3 & 4 \\
MS-AST (EMBC'24)~\cite{zhang2024friends}                 & \checkmark & \textbf{94.5\,$\pm$\,0.0} & 90.4\,$\pm$\,0.2 & 90.0\,$\pm$\,0.4 & 90.1\,$\pm$\,0.2 & -- \\
LoViT (MIA'25)~\cite{liu2025lovit}                       & \checkmark & 92.4\,$\pm$\,6.3 & 89.9\,$\pm$\,6.1 & 90.6\,$\pm$\,4.4 & 90.2\,$\pm$\,6.1 & 4 \\
TUNeS (TBME'25)~\cite{funke2025tunes}                    & \checkmark & 94.2\,$\pm$\,0.6 & -- & -- & -- & 4 \\
KAFR (ours)                                              & \checkmark & 91.5\,$\pm$\,0.4 & \textbf{90.5\,$\pm$\,1.1} & \textbf{91.6\,$\pm$\,0.2} & \textbf{91.0\,$\pm$\,0.6} & \textbf{0.58} \\
\addlinespace
Trans-SVNet    &           & 89.1\,$\pm$\,7.0 & \textbf{84.7\,$\pm$\,7.6} & 83.6\,$\pm$\,6.6 & 84.1\,$\pm$\,7.5 & 4 \\
LoViT          &           & \textbf{91.5\,$\pm$\,6.1} & 83.1\,$\pm$\,9.3 & 86.5\,$\pm$\,5.5 & 84.7\,$\pm$\,9.3 & 4 \\
KAFR baseline (ours)         &           & 88.9\,$\pm$\,0.5 & 83.3\,$\pm$\,2.1 & \textbf{87.5\,$\pm$\,1.0} & \textbf{85.1\,$\pm$\,1.6} & \textbf{1.5} \\
\bottomrule
\end{tabular}
\end{table}

\subsection{Analysis of KAFR under Visual Augmentation}

\begin{figure} [th!]
\centering
\includegraphics[keepaspectratio,width=0.95\textwidth]{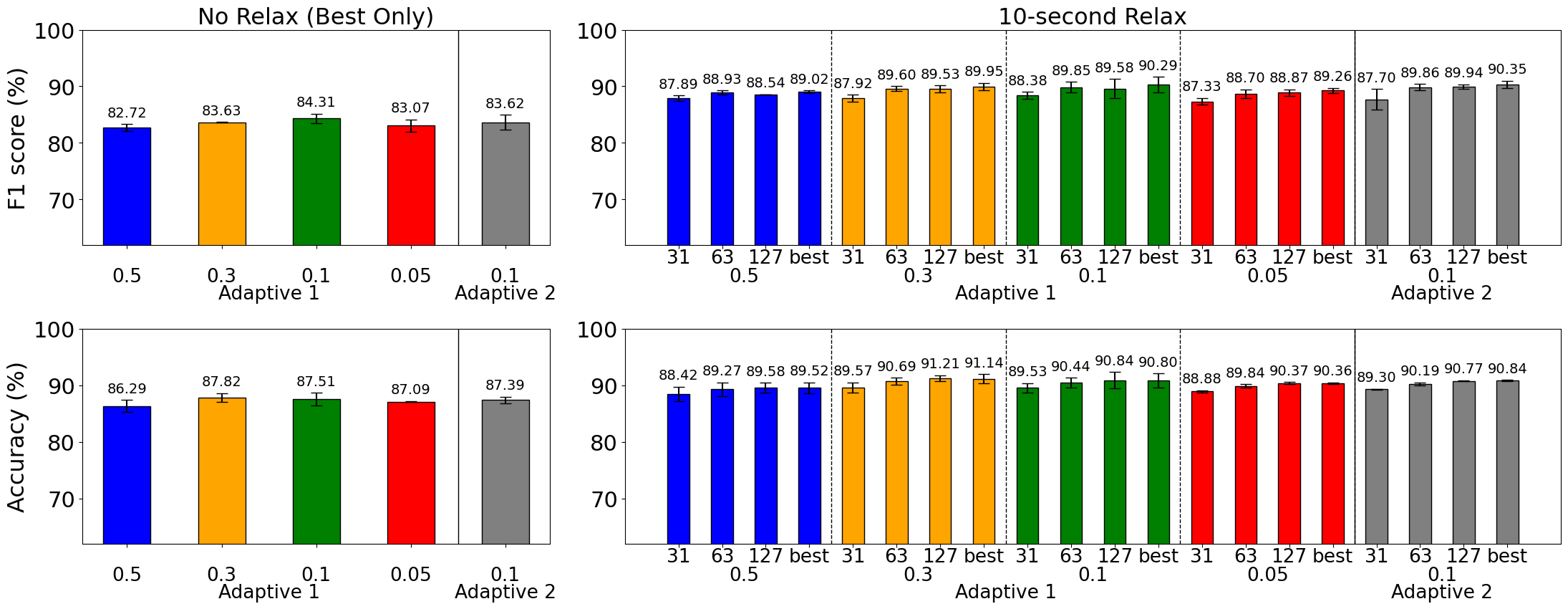}
\caption{Performance of KAFR under visual augmentation. F1 score (top) and accuracy (bottom) across frame retention rates (0.05 to 0.5) and smoothing window sizes ($w = 31, 63, 127$) for Adaptive 1 and Adaptive 2. Left: strict evaluation; right: 10-second relaxed evaluation.}
\label{fig:result_1}
\end{figure}
We conducted experiments in Setting 1 using variations of the proportion of retained frames (0.5, 0.3, 0.1, 0.05) and smoothing window sizes ($w=31, 63, 127$). Due to the large configuration sweep, exploratory grid experiments were performed twice. Figure~\ref{fig:result_1} illustrates the F1 score and accuracy across these settings under both the no-relax (left) and 10-second relax (right) evaluation protocols. Under the no-relax condition, the best F1 score was 84.31\%\,$\pm$\,0.82\% and accuracy was 87.82\%\,$\pm$\,0.77\%, both achieved at portion 0.1 and portion 0.3 using Adaptive 1. This outperformed Adaptive 2 (with the best F1 score of 83.62\%\,$\pm$\,1.34\%). Under the 10-second relax condition, the highest F1 score of 90.35\%\,$\pm$\,0.66\% with accuracy of 90.84\%\,$\pm$\,0.11\% was achieved at \textit{portion 0.1} (10\% of the training data) using Adaptive 2, showing only a 0.69\% decrease in F1 score compared to the full-data baseline (90.98\%). Interestingly, decreasing the portion to 0.05 in Adaptive 1 still yielded strong performance: F1 score of 89.26\%\,$\pm$\,0.45\% and accuracy of 90.36\%\,$\pm$\,0.11\%, despite a 20-fold reduction in data compared to the full data.

\begin{figure} [tb!]
\centering
\includegraphics[keepaspectratio,width=0.85\textwidth]{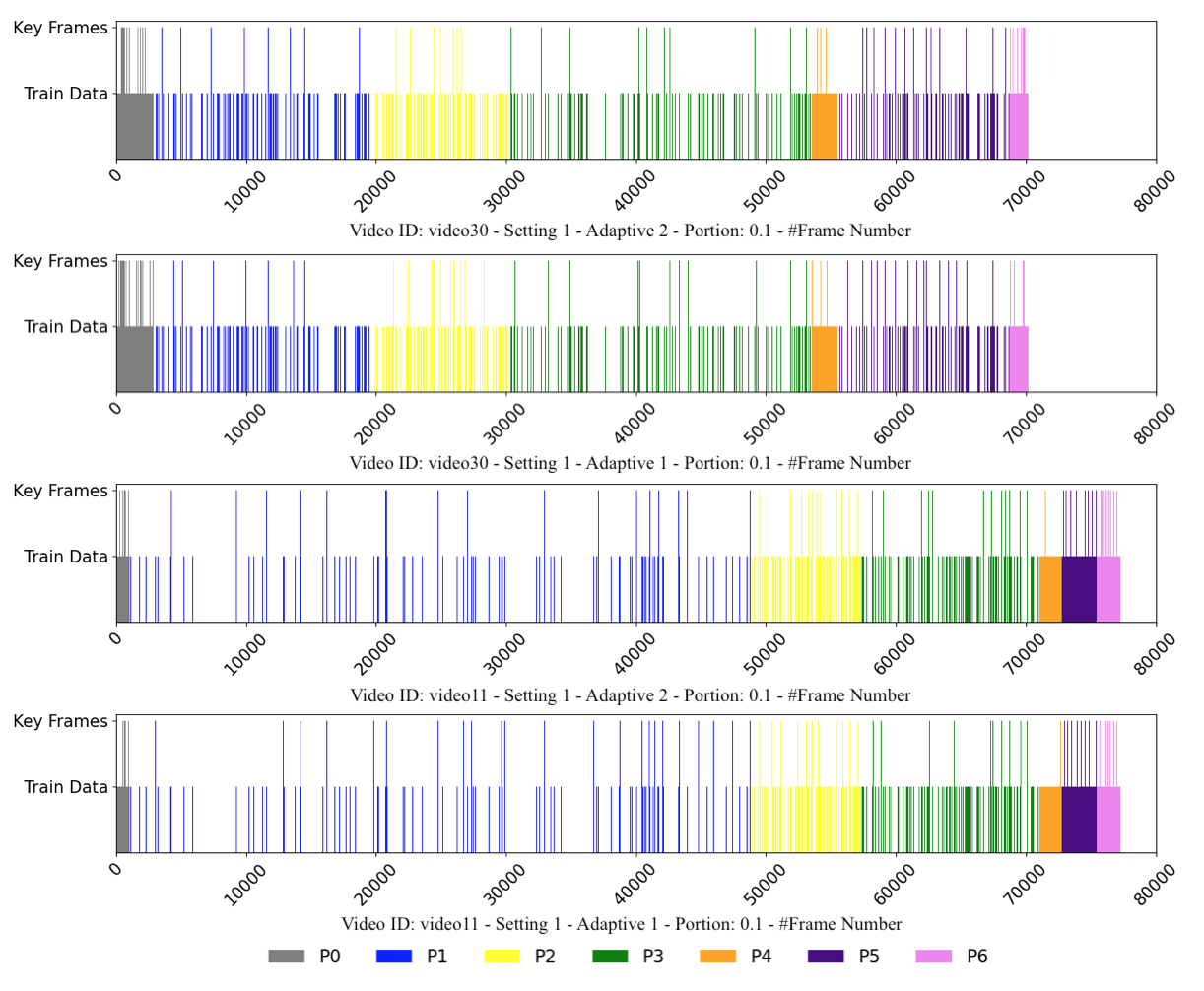}
\caption{Phase segmentation across two sample videos from the Cholec80 dataset. Each subplot displays the annotated surgical phases (P0--P6) using color-coded segments along the video timeline. For each video, the top row represents the key frames retained after applying KAFR, and the bottom row shows the full set of annotated training data.}
\label{fig:segmentation}
\end{figure}

Figure~\ref{fig:segmentation} shows the sampling behavior of Adaptive 1 versus Adaptive 2 on two video samples (video30 and video11). Adaptive 1, using tool velocity, tends to \textit{concentrate sampling within the central regions of each phase}, while avoiding the ambiguous boundaries where phase transitions occur. For example, in video30, Adaptive 1 minimizes dense sampling near the P2–P3 and P4–P5 transitions, focusing instead on frames that reflect stable surgical activity within each phase. In contrast, Adaptive 2, using tool acceleration, tends to \textit{sample more densely around phase transitions}, where abrupt changes in motion might occur. This leads to an increased presence of transition-phase frames, particularly noticeable in video11 near the P1–P2 and P2–P3 boundaries. While this strategy captures dynamic motion cues, it also increases the risk of adding ambiguous or mislabeled frames that may confuse the model during training.

\subsection{Analysis of KAFR under Joint Visual--Temporal Modeling}

\begin{figure} [tbhp!]
\centering
\includegraphics[keepaspectratio,width=0.95\textwidth]{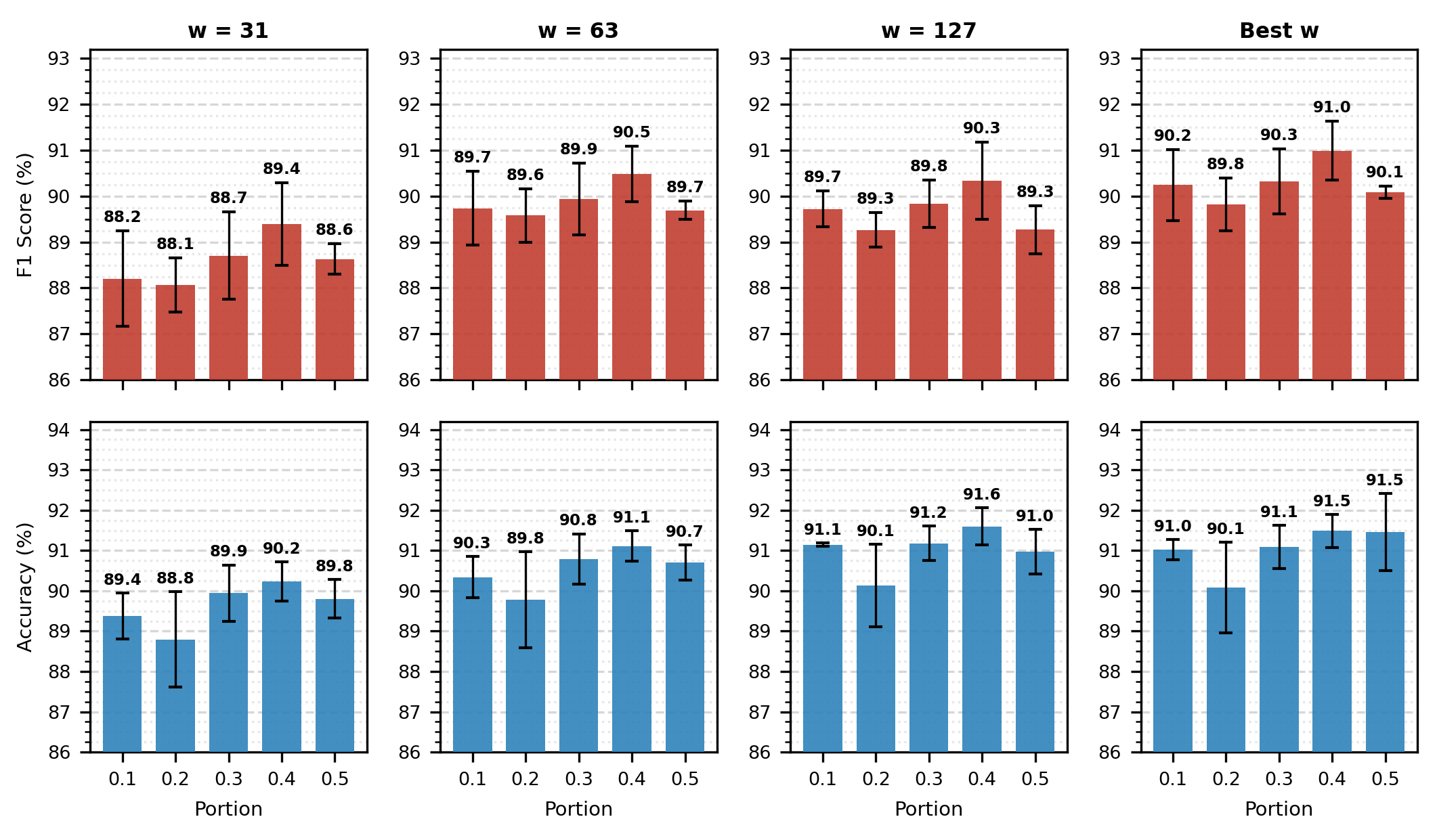}
\caption{Performance of KAFR under joint visual and temporal modeling. Evaluation of accuracy (left) and F1 score (right) across multiple frame portions and smoothing window sizes, incorporating both visual augmentation and kinematics filtering. The best values are obtained by searching over window sizes ranging from 2 to 200 (post-hoc sensitivity analysis on the test set).}
\label{fig:result_3}
\end{figure}

In this experiment, we use only Adaptive 1, as it consistently outperforms Adaptive 2 in phase segmentation performance. In addition, combining Adaptive 1 and Adaptive 2 would result in a substantial increase in the number of experimental configurations, so we restrict our evaluation to Adaptive 1 for both effectiveness and practicality. In Setting 2, a filter module was added to address the impact of camera movement on tool-motion signals. The SMA window size was varied from 2 to 200 frames on the test-set predictions as a post-hoc sensitivity analysis to examine the effect of smoothing-window length, rather than as a training-time hyperparameter optimization. As shown in Figure~\ref{fig:result_3}, the results indicate that using a frame portion of 0.4 consistently yields the best performance across window sizes. Specifically, the highest F1 score of 91.0\%\,$\pm$\,0.6\% and accuracy of 91.5\%\,$\pm$\,0.4\% are observed at portion 0.4 under the most favorable smoothing-window setting in this sensitivity analysis.
\begin{figure} [tbhp!]
\centering
\includegraphics[keepaspectratio,width=0.85\textwidth]{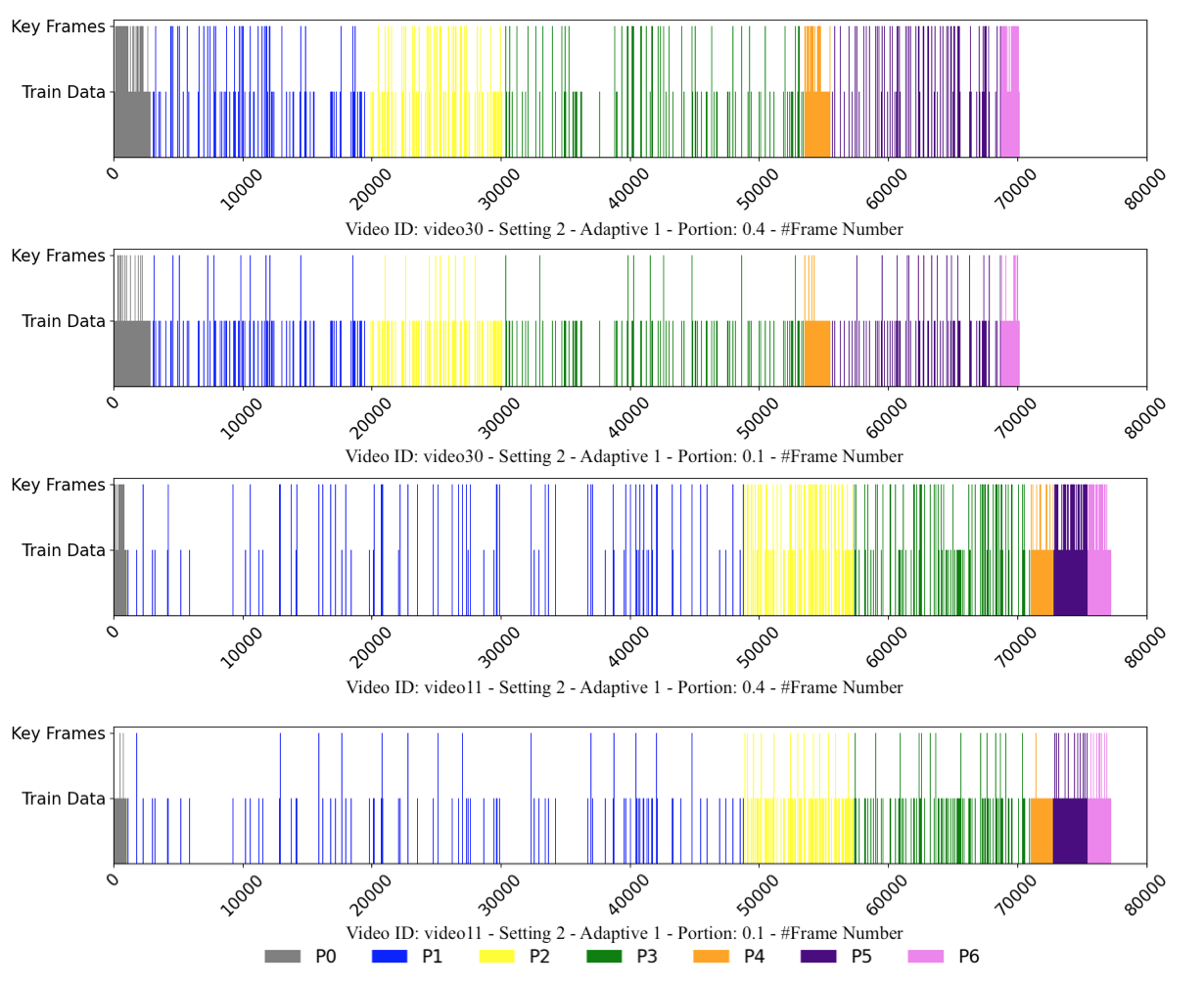}
\caption{Phase segmentation across two sample videos from the Cholec80 dataset. Each subplot displays the annotated surgical phases (P0--P6) using color-coded segments along the video timeline. For each video, the top row represents the key frames retained after applying KAFR, and the bottom row shows the full set of annotated training data.}
\label{fig:segmentation2}
\end{figure}
\begin{figure} [tbp!]
\centering
\includegraphics[keepaspectratio,width=0.95\textwidth]{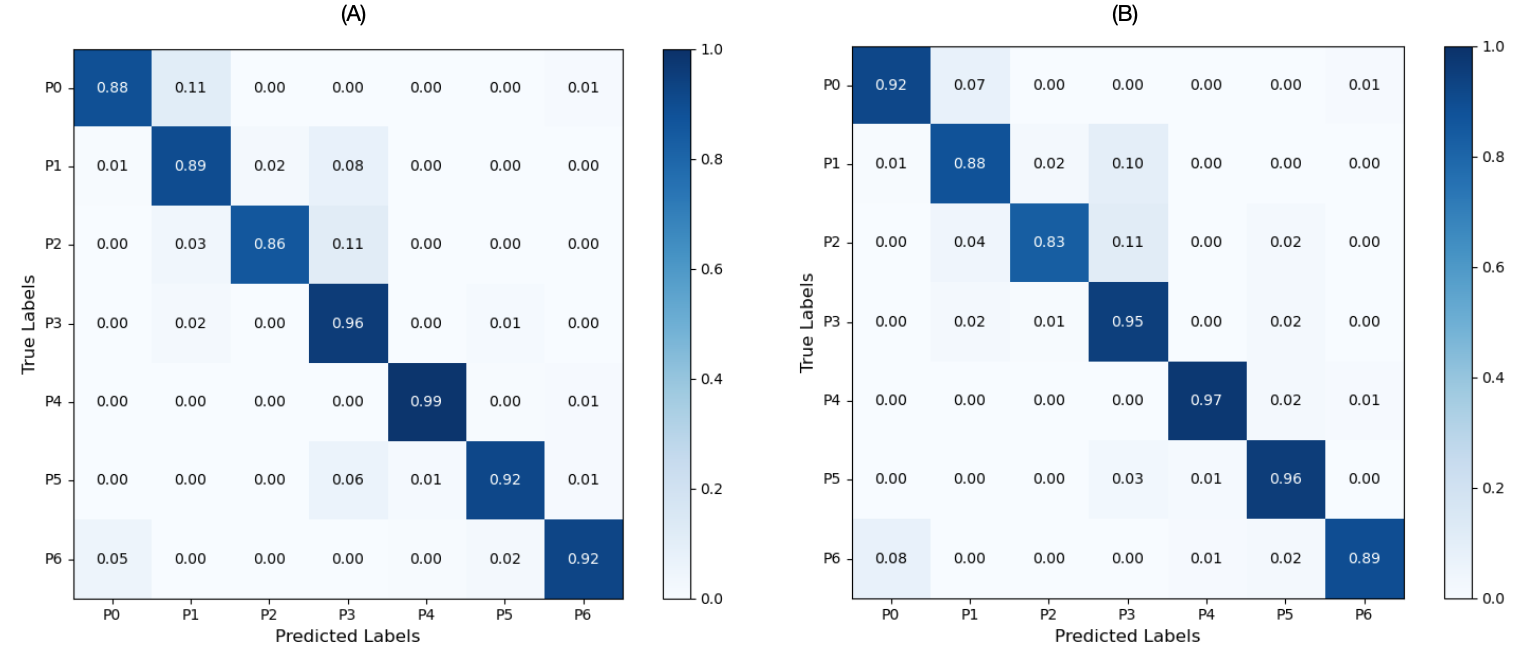}
\caption{Normalized confusion matrices. (A) Adaptive 1 in Setting 2 with frame portion 0.4. (B) Adaptive 2 in Setting 1 with frame portion 0.1.}
\label{fig:confusion_matrix}
\end{figure}

Figure~\ref{fig:segmentation2} shows that Adaptive 1 maintains consistent sampling behavior across two video samples at two portion settings (0.1 and 0.4). At portion 0.4, particularly in video30, we can see dense and evenly spaced key frames within the central segments of phases such as P2 (yellow) and P3 (green), but transition zones between phases, for instance, between P1 and P2 or P4 and P5, are sparsely sampled. A similar pattern can be seen in video11, where frames are concentrated in the middle of stable phases like P1 and P3, while the regions near transitions into P2 and P5 are minimally represented. This consistent sampling pattern across videos and reduction levels indicates that Adaptive 1 systematically prioritizes motion stability, avoiding ambiguous transition phases.

\subsection{Confusion Analysis Across Settings}
\label{sec:confusion_matrix}

Figure~\ref{fig:confusion_matrix} illustrates the normalized confusion matrices obtained from the best-performing models under two distinct setting scenarios. In both matrices, the model demonstrates strong overall classification performance, especially for phases P3, P4, and P5, where prediction accuracy exceeds 92\%. However, notable confusion occurs in certain transitions. For instance, in subfigure (A), phase P1 is occasionally misclassified as P3 (8\%), and P2 is misclassified as P3 (11\%), indicating that the model struggles to distinguish between these mid-procedure phases, likely due to similar tool motions. In addition, confusion between P0 and P1 (11\%) may stem from ambiguous early procedural frames. In subfigure (B), Adaptive 2’s sampling showed slightly improved performance in the early phase, with P0 reaching 92\%. However, misclassifications between P2 and P3 (11\%) and between P1 and P3 (10\%) still persist, while performance decreased in the final phase, with P6 reduced to 89\%.

\subsection{Model Interpretability and Clinical Insight}

\begin{figure} [tbp!]
\centering
\includegraphics[keepaspectratio,width=0.85\textwidth]{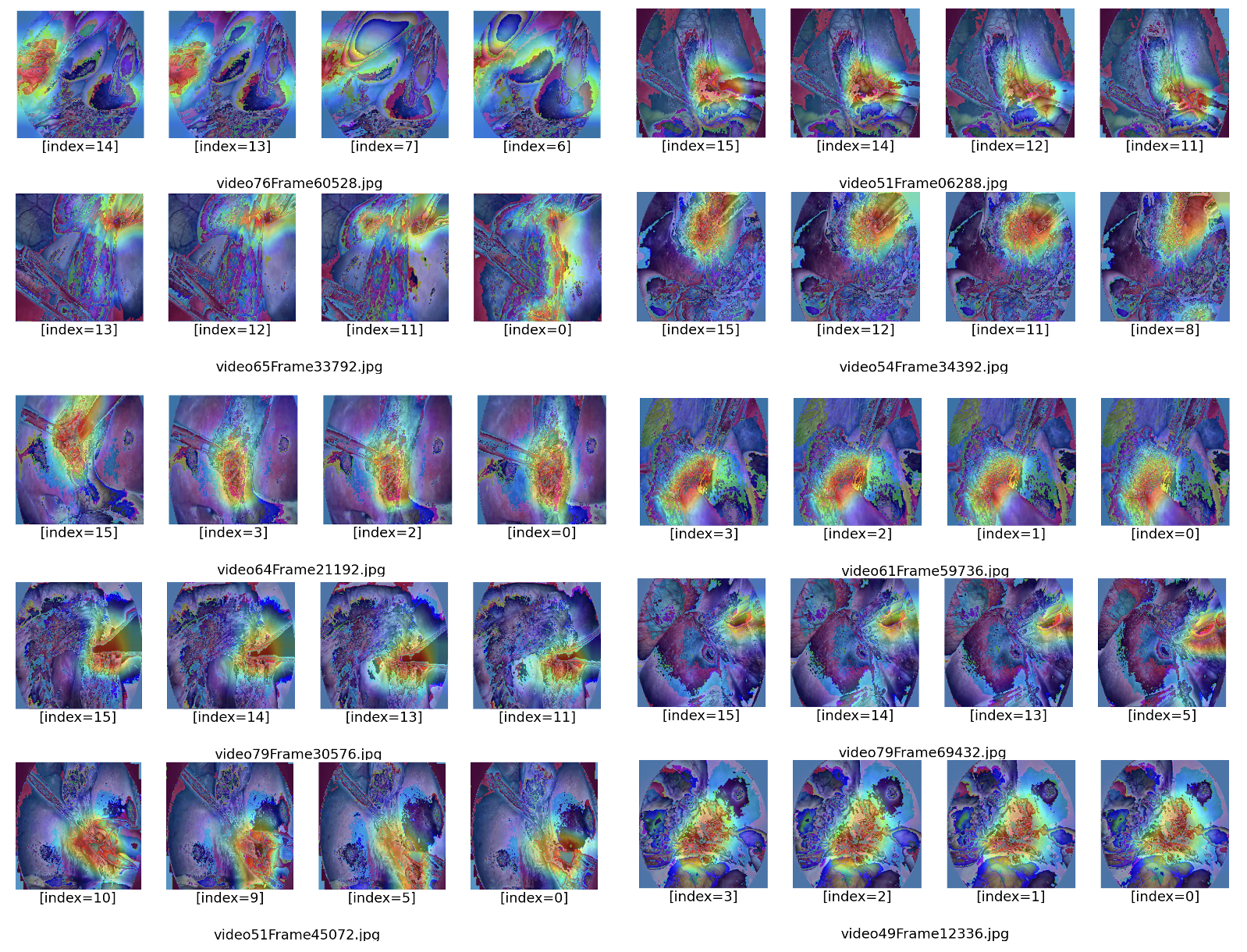}
\caption{Grad-CAM visualizations highlighting model attention across selected frames from two samples. The four most informative frames, ranked by Grad-CAM activation intensity, are displayed from a 16-frame clip. Index values denote the relative position of each frame (counting backward) within the clip.}
\label{fig:grad_cam}
\end{figure}

To support interpretability, we used Grad-CAM visualizations to identify where the model focuses during prediction. The heatmaps in Figure~\ref{fig:grad_cam} illustrate that the model consistently focuses on central regions of the surgical field, where instruments interact with anatomical structures. The most activated regions (shown in red/yellow) correspond to tool-tissue contact zones, instrument tips, or anatomical landmarks, suggesting that the model attends to regions of surgical activity rather than background features.

The attention maps also demonstrated temporal consistency, with the model's focus remaining localized across adjacent frames. Minimal activation at the image periphery indicates robustness against background distractions.

%
\section{Discussion}
\label{sec:discussion}
This study presents KAFR, a framework for surgical phase segmentation that selects frames based on kinematic saliency rather than fixed temporal intervals. We successfully extended the approach from robotic surgery~\cite{nguyen2025kinematic} to the more visually complex domain of laparoscopic cholecystectomy. Our findings demonstrate that accurate phase segmentation does not require dense temporal sampling; rather, it can be achieved by analyzing a sparse subset of frames characterized by substantial tool motion, reducing data requirements approximately 7-fold compared to existing methods while maintaining competitive accuracy.

The motion-based frame selection strategy in KAFR resembles the selective attention patterns of experienced surgeons. During an operation, surgeons do not attend equally to every moment; their concentration is naturally drawn to periods of active tissue manipulation rather than tool repositioning or waiting. Similarly, KAFR preferentially selects frames with high kinematic activity. Our finding that Adaptive 1 (displacement-based selection) focuses sampling within stable phases while avoiding confusing transitions aligns with the attention pattern of frames capturing consistent, intentional tool movement are more informative for phase segmentation.

This pattern is also reflected in the distribution of errors. The confusion matrix analysis (Figure~\ref{fig:confusion_matrix}) demonstrated that misclassifications occurred around clinically similar phases, particularly P1 (Calot triangle dissection) and P3 (Gallbladder dissection). This pattern reflects inherent procedural ambiguity rather than model deficiency: both phases involve similar tools performing dissection on adjacent anatomical structures, and the visible boundary between them is often subjective, even among expert observers. That the model encounters difficulty in the locations where human annotators differ indicates it is learning meaningful surgical features rather than accidental visual patterns.

Aside from accuracy, the pipeline’s computation cost is something to consider. To provide comprehensive efficiency accounting, we analyzed the computational expenses of all pipeline components separately. The KAFR pipeline involves three stages: (i) YOLOv8 instrument detection, which runs at the full native frame rate during preprocessing and requires approximately 15~ms per frame; (ii) adaptive frame selection, a lightweight step requiring less than 1~ms per frame to compute kinematic saliency scores from tool centroids; and (iii) X3D phase classification on the retained frames only. On our hardware (two NVIDIA RTX A6000 GPUs), per-epoch training time for full data was approximately 23 minutes 42 seconds, while training with 10\% frame retention required 8 minutes 12 seconds per epoch, a 2.9-fold per-epoch speedup. The X3D-M phase classifier requires 10.12~GFLOPs per 16-frame input clip and has 3.79~M parameters, measured using the thop profiler on our hardware. At inference, X3D-M achieves a throughput of approximately 710~frames per second (22.5~ms per 16-frame clip) on an NVIDIA RTX A6000 GPU. Because KAFR retains only 0.58\% of frames for phase classification, the effective X3D inference load per video is reduced by the same factor, yielding a proportional reduction in total X3D FLOPs processed per video. The 0.58\% retained-frame figure refers specifically to the phase classification module; the tool detector runs at the full native frame rate, and consequently the end-to-end pipeline does not achieve a 7-fold reduction in total inference time.

The internal behavior of the classifier is consistent with these efficiency findings. The Grad-CAM visualizations showed that the model prioritizes tool-tissue interaction zones above background features (Figure~\ref{fig:grad_cam}). These findings align with attention-based approaches like Rendezvous~\cite{NWOYE2022rendezvous}, which demonstrated that detecting surgical activities relies largely on capturing the spatiotemporal interplay between tools and their anatomical targets. The observed localized focus, combined with temporal consistency between subsequent frames, offers insight into model behavior.

These properties indicate that KAFR generalizes across imaging modalities. The successful adaptation of KAFR from robotic to laparoscopic surgery demonstrates that motion-based frame selection is robust to imaging conditions. Robotic surgery produces steady, high-resolution video with minimal camera motion, while laparoscopic surgery involves handheld camera control, variable image quality, and frequent motion artifacts. KAFR’s competitive performance in these distinct settings suggests it is not overfitted to specific capture modalities. This indicates strong potential for domain transferability to other video-based tasks, where relevant content is sparse, such as gastrointestinal endoscopy or even non-medical domains like surveillance video summary. Future work will focus on validating KAFR on these various datasets to further establish its applicability as a domain-agnostic frame selection tool.

Beyond the present study, KAFR has been validated on two additional robotic surgical datasets from distinct referral centers~\cite{nguyen2025kinematic}: (1) Gastrojejunostomy (GJ, 42~videos, 6~phases), where KAFR achieved a tenfold frame reduction with an accuracy improvement of 4.32\% and an F1 gain of 0.16\%; and (2) Pancreaticojejunostomy (PJ, 100~videos, 6~phases), where KAFR achieved a fivefold frame reduction with an accuracy improvement of 2.05\% and an F1 gain of 2.54\%. These results were achieved at two different referral centers with robotic surgical systems, representing a different modality and patient population than the laparoscopic Cholec80 benchmark evaluated herein.

Furthermore, the YOLOv8 detection component trained on Cholec80 was subsequently applied without re-annotation to detect graspers in fundoplication and crural repair procedures, confirming transferability of the detection component across procedure types~\cite{nguyen2025kinematic}. Taken together, KAFR has now been validated across three independent datasets, two surgical modalities (robotic and laparoscopic), two procedure types, and two institutions, covering over 220 videos. The present work therefore constitutes the third independent validation of the KAFR framework, extending it to the challenging laparoscopic domain for the first time.

X3D was selected as the phase classification backbone for three main reasons: (1) it provides a strong balance between spatiotemporal modeling capacity and computational efficiency, requiring only {\raise.17ex\hbox{$\scriptstyle\sim$}}10.12~GFLOPs per 16-frame clip with 3.79~M parameters and achieving approximately 710~frames/second on an NVIDIA RTX A6000 GPU; (2) it is well suited for clip-based inference on the trimmed frame sequences selected by KAFR; and (3) it allows direct comparison with our prior validation work in robotic surgery~\cite{nguyen2025kinematic}. The MS-TCN family (TeCNO~\cite{czempiel2020tecno}) is already included in Table~\ref{tab:comparison} as a primary baseline. TUNeS~\cite{funke2025tunes} has been added to Table~\ref{tab:comparison}, reporting 94.2\%\,$\pm$\,0.6\% accuracy with 4\% retained frames in the offline setting. Importantly, KAFR is classifier-agnostic: the frame selection component is not coupled with the downstream classification model, and combining KAFR with MS-TCN, TUNeS, or any other temporal classification architecture is a promising direction for future work.

There are some limitations of this work that need to be recognized. First, the present study evaluates KAFR on Cholec80, a single-center dataset of one type of laparoscopic procedure. We note that KAFR has been validated in prior work on two additional robotic surgical datasets from two distinct referral centers: Gastrojejunostomy (GJ, 42~videos) and Pancreaticojejunostomy (PJ, 100~videos), totaling over 220 videos in total across two modalities and two institutions~\cite{nguyen2025kinematic}. Nevertheless, additional validation in multiple centers and assessment in other types of laparoscopic procedures would add strength to the claims of generalizability. Furthermore, the performance comparison with state-of-the-art methods is currently restricted to Cholec80; whether KAFR’s performance margins generalize to other benchmarks remains to be evaluated.

Secondly, the current implementation relies on annotated data to fine-tune the tool detector. Although this strategy enables accurate tracking and allows the trained detector to operate automatically during inference, the annotation process requires upfront effort: for the Cholec80 experiments, the segmentation mask annotation required approximately 1 week of a dedicated research assistant time.
When the model was developed, promptable segmentation foundation models such as SAM were not yet available; future adaptations could leverage SAM-based annotation tools or pre-trained surgical tool detectors to significantly reduce this annotation burden~\cite{JASPERS2026scaling,ross2018exploiting}.

Thirdly, visual occlusions caused by surgical smoke, blood, or lens fogging can significantly limit tool detection accuracy. KAFR circumvents this reliance on visual quality by prioritizing kinematic activity. When adverse events like bleeding occur, the surgeon’s immediate response, such as rapid suctioning or applying pressure, generates significant tool motion. KAFR captures these high-activity sequences, effectively tracking the surgical intervention.

Fourthly, hyperparameters were selected empirically rather than through systematic optimization, and computational savings did not scale linearly with frame reduction due to fixed processing overheads.

Finally, the utility of retroactive video analysis for applications such as surgical training, quality review, and video indexing was assessed in this study. Real-time intraoperative monitoring represents a unique application with different design requirements, and clinical utility in such workflows remains to be validated.

In summary, KAFR achieves an F1 score of 91.0\% on the Cholec80 benchmark, comparable to state-of-the-art methods, utilizing only 0.58\% of frames for phase-classification training, thereby reducing data requirements by 7-fold. Building on prior validation on Gastrojejunostomy and Pancreaticojejunostomy datasets~\cite{nguyen2025kinematic} with over 220 videos from two institutions and two surgical modalities, KAFR has been independently validated on three datasets. The successful generalization from robotic to laparoscopic surgery, despite significant differences in image quality and camera stability, demonstrates that motion-based frame selection generalizes across surgical scenarios and underlines its potential as a domain-agnostic preprocessing approach for efficient surgical video analysis.
\section{Methods}
\label{sec:methods}
\subsection{Architecture}
\begin{figure} [bt!]
\centering
\includegraphics[keepaspectratio,width=0.95\textwidth]{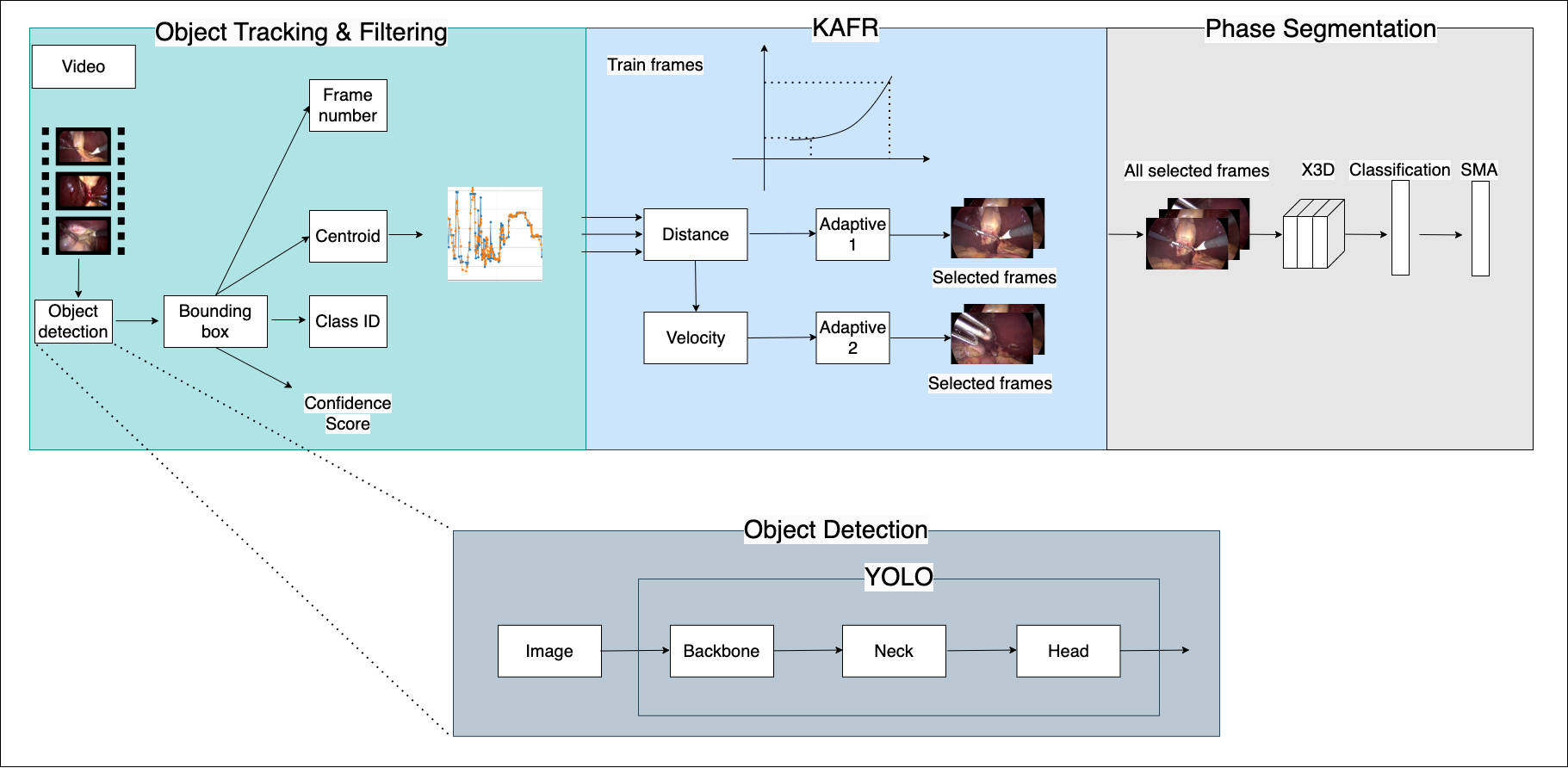}
\caption{Overview of the KAFR architecture. The pipeline involves three phases: (1) Object Tracking and Filtering detects and tracks surgical tools and filters movements; (2) KAFR computes centroids and identifies critical frames using tool movement; and (3) Phase Segmentation classifies frames into segments with X3D CNN and SMA. KAFR, Kinematics Adaptive Frame Recognition; CNN, Convolutional Neural Network; SMA, Smooth Moving Average.}
\label{fig:architecture}
\end{figure}

Our proposed KAFR framework includes three essential stages: Object Tracking and Filtering, KAFR, and Phase Segmentation (Figure~\ref{fig:architecture}). In the first stage, Object Tracking and Filtering is essential for detecting surgical instruments in noisy visual scenes characterized by occlusion, motion blur, and cluttered surgical fields. This module extracts bounding boxes, class IDs of the surgical tools, and corresponding frame numbers from the input video. Within the stage, we compute the centroids of each detected tool to trace their movement across frames (limited to those in the training set). Since tools are detected as polygons, their centroids are computed. The results are then filtered through Infinite Impulse Response (IIR) Butterworth filter to minimize insignificant variations and background noise. In the second stage, KAFR utilizes tool displacement and velocity variation to identify frames associated with significant tool motion. In the third stage, we apply an X3D CNN model to classify these frames into distinct surgical phases and Smooth Moving Average (SMA) to reduce misclassified frames.
\subsection{Adaptive 1 - Velocity Based}

\begin{figure} [tbh!]
\centering
\includegraphics[keepaspectratio,width=0.95\textwidth]{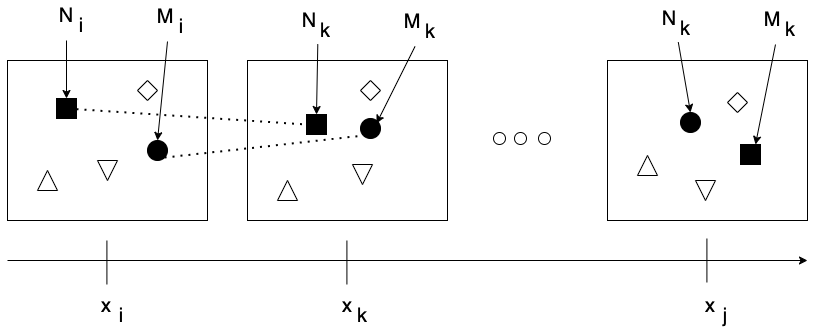}
\caption{An illustration of Adaptive Frame Recognition. The subset $s$ includes two tools, $M$ and $N$ (depicted as circle and square shapes), which appear within a cluttered surgical field that may comprise abdominal organs, additional instruments, and visual noise (represented by various other shapes). These tools are continuously tracked across video frames, with the frame index $k$ advancing consecutively until a predefined threshold $d$ is achieved.}
\label{fig:proposed}
\end{figure}
Within a video sequence of $n$ frames $\{x_1, x_2, ..., x_n\}$, a threshold $d$ determines the distance measure $D(x_i, x_j)$ used to classify a pair of frames $(x_i, x_j)$ as key frames or similar frames. A pair is classified as a key frame pair, denoted $P_{\text{key}}$, if the distance between $x_i$ and $x_j$ is less than or equal to $d$. The frames lying between two such key frames are considered similar and are denoted as $P_{\text{similar}}$. Importantly, this selection is performed at the video level rather than at the clip or phase level, meaning the search for key frame pairs proceeds sequentially across all frames of the entire video and the adaptive threshold is applied globally. We define the set of all key frame pairs $(K)$ as follows (Figure~\ref{fig:proposed}):
\begin{equation}
K(d) = \{(x_i,x_j) \mid D(x_i,x_j) \leq d, (x_i,x_j) \in P_{key}\},
\label{eq:f1}
\end{equation}
Let $S$ be a set of points representing pixels in a frame, and let $s \subseteq S$ be a subset of interest. The distance between frames $x_i$ and $x_j$, denoted $D(x_i, x_j)$, is defined as follows:
\begin{equation}
D(x_i, x_j) = f\left(\sum\limits_{s \in S}\sum_{k=i+1}^{j}\left|\left|s(x_i)-s(x_k)\right|\right|\right),
\end{equation}
where $\left|\left|\cdot\right|\right|$ being the Euclidean norm and $f\,:\,\mathbb{R}\rightarrow \mathbb{R}$ being a decreasing (or at least non-increasing) function. In \underline{Adaptive 1}, we assume that
\begin{equation}
f(z_d)= \frac{1}{\left(z_d+\epsilon\right)^{\beta_d}},
\end{equation}
where constant $\beta_d>0$ and $\epsilon$ is a small number introduced to avoid division by zero.

\subsection{Adaptive 2 - Acceleration Based}

An alternative to the equation~\eqref{eq:f1} is to use variation of \emph{velocity} rather than distance
\begin{equation}
K(d) = \{(x_i,x_j) \mid V(x_i,x_j) \leq d, (x_i,x_j) \in P_{key}\},
\end{equation}

Assuming $S$ is a set of points of a frame and $s$ is a subset $S$, the variation of velocity $V(x_i, x_j)$ is defined as
\begin{equation}
V(x_i, x_j) = f\left( \sum\limits_{\substack{s \in S}} \sum_{k=i+1}^{j} \left| V_{\text{s}}(x_i) - V_{\text{s}}(x_k) \right| \right),
\end{equation}
where $V_{\text{s}}(x_k)$ is the velocity of point {$x_k$} in the subset $s$.

In \underline{Adaptive 2}, it is assumed that
\begin{equation}
f(z_v)= \frac{1}{\left(z_v+\epsilon\right)^{\beta_v}}.
\end{equation}
where constant $\beta_v>0$ and $\epsilon$ is a small number introduced to avoid division by zero.

\subsection{Phase Segmentation}
To classify surgical phases from laparoscopic videos, we utilized the X3D architecture~\cite{feichtenhofer2020x3d}, an efficient 3-D CNN model optimized for spatiotemporal learning. Our approach incorporates several advanced enhancements to improve accuracy, robustness, and temporal coherence. The details of these techniques are described in the following sections.

For phase classification, we employ the X3D architecture~\cite{feichtenhofer2020x3d}, a 3-D convolutional neural network designed to balance recognition accuracy and computational efficiency. Unlike standard 3-D CNNs that apply uniform scaling across all dimensions, X3D progressively expands the network along multiple axes, including temporal extent, spatial resolution, channel width, and 
network depth, achieving strong video classification performance with reduced computational cost. The architecture employs inverted bottleneck blocks~\cite{sandler2018mobilenetv2}, which expand features internally before projecting back to lower dimensions, providing parameter efficiency 
suitable for processing the high volume of surgical video data. The X3D model was fine-tuned from a pre-trained checkpoint trained on the Kinetics-400 dataset.

Our approach differs from previous approaches~\cite{al2024development}, which select frames uniformly at predetermined intervals irrespective of content. Each input sequence consists of a desired number of consecutive KAFR-selected frames, which the X3D model processes to provide a probability distribution over the seven surgical phases defined in Cholec80.

Since the Cholec80 dataset is unbalanced, with surgical phases varying greatly in duration, we apply a resampling strategy to normalize frame counts across phases~\cite{al2024development}.  
Let \( d_{ij} \) denote the duration (in frames) of phase \( j \in \{1,\dots,t\} \) in video \( i \in \{1,\dots,n\} \), and define the set of durations for each phase as $D_j = \{ d_{1j}, d_{2j}, \dots, d_{nj} \}$. The per-phase median is $d_{\text{med}}^{(j)} = \text{median}(D_j)$, and the global reference duration is $d_{\text{med}} = \text{median} \big( \{ d_{\text{med}}^{(1)}, \dots, d_{\text{med}}^{(t)} \} \big)$.

Given \( N_{ij} \) frames for phase \( j \) in video \( i \), the resampled number is
\begin{equation}
N_{ij}^{\text{new}} =
\begin{cases}
\left\lceil \dfrac{d_{\text{med}}}{d_{ij}} \, N_{ij} \right\rceil & d_{ij} < d_{\text{med}} \quad \text{(upsampling)} \\[6pt]
\left\lfloor \dfrac{d_{\text{med}}}{d_{ij}} \, N_{ij} \right\rfloor & d_{ij} > d_{\text{med}} \quad \text{(downsampling)}
\end{cases}.
\end{equation}

Frames are duplicated when \( d_{ij} < d_{\text{med}} \) and randomly discarded when \( d_{ij} > d_{\text{med}} \).

The total loss combines cross-entropy and Earth Mover’s Distance (EMD)~\cite{rubner2000earth,al2024development}:
\begin{align}
\mathcal{L}_{\text{total}} &= \mathcal{L}_{\text{CE}} + \lambda \cdot \mathcal{L}_{\text{EMD}}, \\
\mathcal{L}_{\text{CE}} &= - \sum_{c=1}^{C} y_c \log(p_c), \\
\mathcal{L}_{\text{EMD}} &= \frac{ \sum_{i=1}^{m} \sum_{j=1}^{n} f_{ij} \cdot d(p_i, q_j) }{ \sum_{i=1}^{m} \sum_{j=1}^{n} f_{ij} },
\end{align}
where $\lambda$ balances the two loss components, \( p = \{p_1, p_2, \dots, p_m\} \) is the predicted probability distribution, and \( q = \{q_1, q_2, \dots, q_n\} \) is the ground-truth distribution, \( f_{ij} \) denotes the optimal flow from \( p_i \) to \( q_j \), and \( d(p_i, q_j) \) is a ground distance between classes \( i \) and \( j \), typically computed using Euclidean distance.

As a post-processing step to reduce misclassification~\cite{al2024development}, a Smooth Moving Average (SMA) technique is applied:
\begin{equation}
y_t^* = \text{mode}( \hat{y}_{t-k}, \dots, \hat{y}_{t+k} ), \quad \text{with } w = 2k + 1.
\end{equation}
where $y_t^*$ is the smoothed label at time $t$, $\hat{y}_t$ is the predicted label, and $w$ is the window size. The function $\text{mode}(\cdot)$ returns the most frequently occurring label within the window.
\subsection{Object Detection and Filtering}
\label{sec:objectdetection}
\begin{figure} [bt!]
\centering
\includegraphics[keepaspectratio,width=0.85\textwidth]{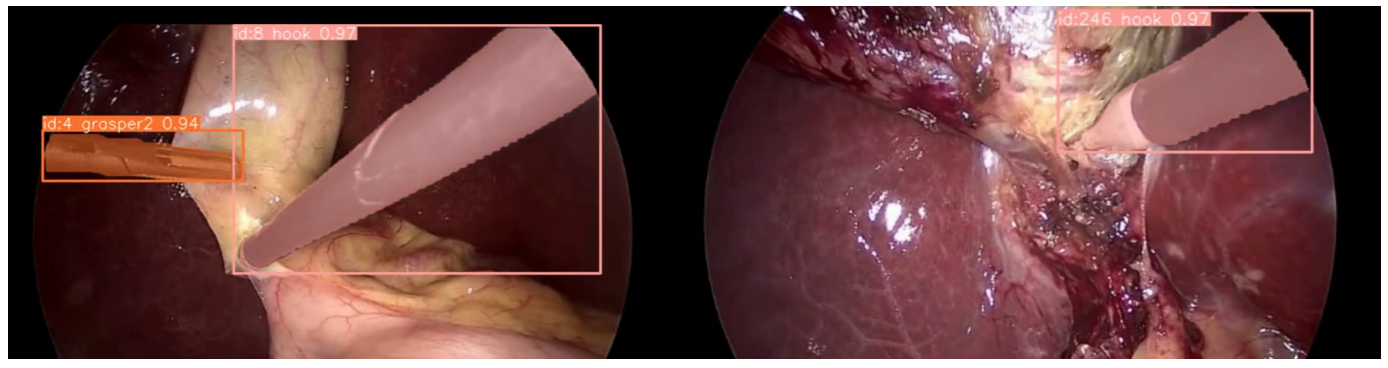}
\caption{Automatic detection of surgical tools using the YOLOv8 model.}
\label{fig:yolo}
\end{figure}
The precise location of tools is essential to our proposed method, which is based on tool presence and motion for downstream analysis. To accomplish this, we utilize You Only Look Once (YOLO) version 8~\cite{redmon2016you,yolov8}, a real-time detection framework offering high accuracy and computational efficiency. Although newer YOLO variants exist, we adopted YOLOv8 as it remains widely used in surgical analysis and our contribution is distinct from the specific detector choice. While initially trained on general-purpose datasets such as COCO~\cite{lin2014microsoft} and ImageNet~\cite{krizhevsky2012imagenet}, the model is fine-tuned on the Cholec80 dataset to ensure reliable detection of surgical instruments. Since Cholec80 offers only binary tool presence labels, tool segmentation was manually annotated by medical experts for a subset of frames to enable supervised fine-tuning and spatial localization. The YOLOv8 architecture is composed of three main components: a feature extraction backbone, a feature aggregation neck, and a prediction head.

The confidence score for class $k$ is computed as:
\begin{equation}
\text{Conf}_{ijk}^{(b)} = \hat{p} \cdot \hat{c}_k \cdot \text{IOU}\left(\hat{b}_{ij}^{(b)}, b_{\text{gt}}\right),
\end{equation}
where the superscript $(b)$ denotes the $b$-th bounding box predicted at spatial location $(i,j)$, with $b = 1, \dots, B$; $\hat{p}$ is the objectness score, which indicates the probability that the bounding box contains an object; and $\hat{c}_k$ is the predicted probability for class $k$. The Intersection over Union (IoU) between the predicted and ground truth bounding boxes is defined as:
\begin{equation}
\text{IOU}(\hat{b}, b_{\text{gt}}) = \frac{\text{area}(\hat{b} \cap b_{\text{gt}})}{\text{area}(\hat{b} \cup b_{\text{gt}})},
\end{equation}
%
In our setup, only detections with confidence scores above $0.5$ are retained for subsequent centroid tracking and frame selection.

A portion of the Cholec80 video dataset was utilized to train the YOLOv8 model. Although the same dataset supports both surgical tool detection and phase segmentation, these are distinct tasks with different goals, and their use in parallel should not compromise each other’s performance. Since YOLOv8 outputs tools as polygons, their centroids are computed using the GeoPandas library in PyTorch. Figure~\ref{fig:yolo} illustrates the use of YOLO for detecting and segmenting surgical tools.

The tool motion is then filtered to retain precise movement. This filtering process is modeled in discrete time using an IIR Butterworth low-pass filter, characterized by the filter order \( n \) and the normalized cutoff frequency \( W_n = \frac{2f_c}{f_s} \), where \( f_s \) is the sampling frequency and \( f_c \) is the cutoff frequency. In our experiments, we use the \texttt{SciPy} library to compute the filter coefficients and apply zero-phase filtering to cancel out phase shift by applying forward and backward filtering.

\begin{figure} [bt!]
\centering
\includegraphics[keepaspectratio,width=0.85\textwidth]{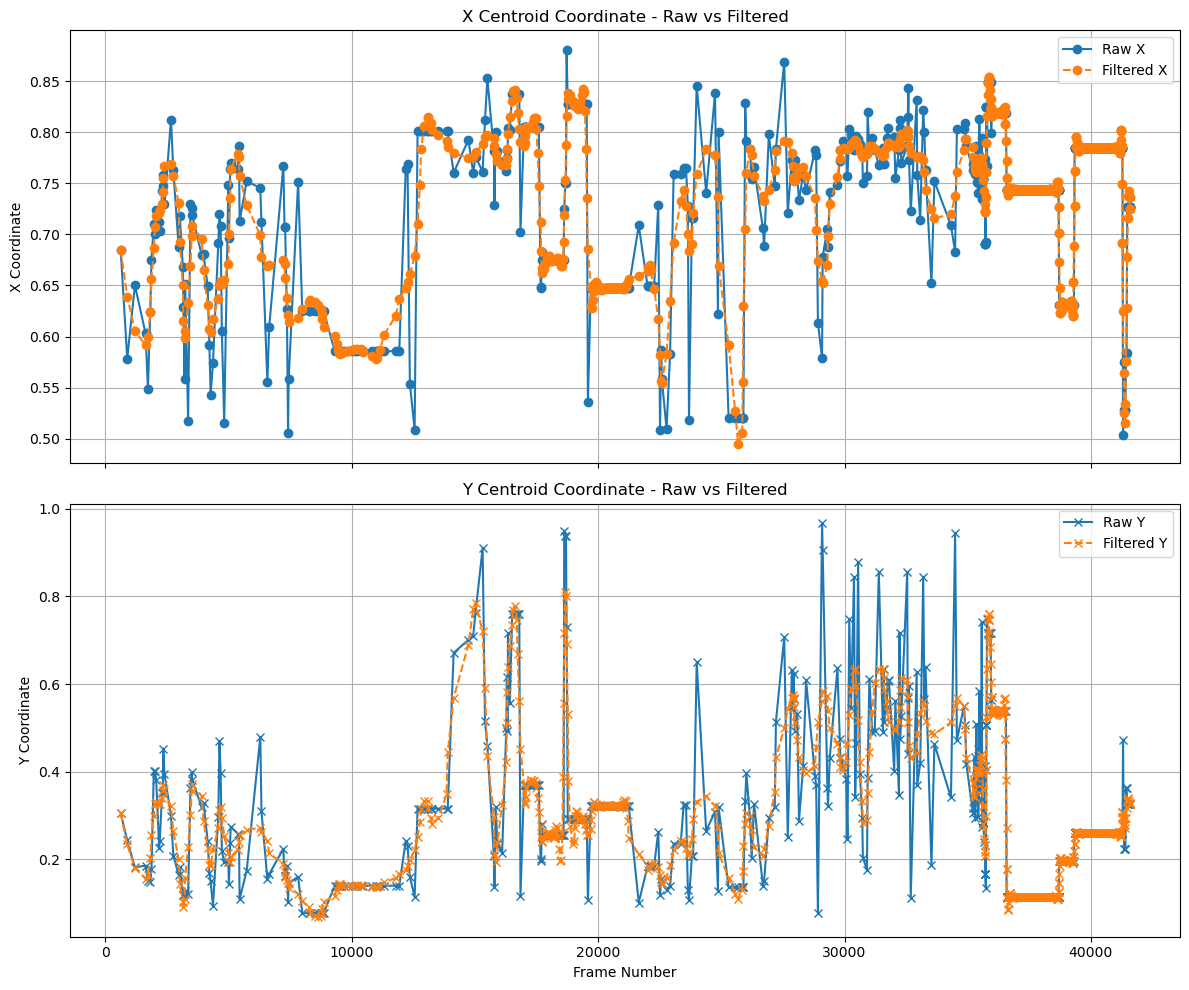}
\caption{Comparison of raw and filtered centroid trajectories for a surgical tool across video frames. The top plot displays the X-coordinate, while the bottom plot shows the Y-coordinate. The raw coordinates (blue) exhibit high-frequency noise and abrupt fluctuations, but the filtered trajectories (orange) show smoother and steadier motion. An IIR Butterworth filter was applied to emphasize deliberate tool movement and suppress background jitter. IIR, Infinite Impulse Response.}
\label{fig:filter}
\end{figure}

Figure~\ref{fig:filter} shows an example of raw versus filtered centroid trajectories. The raw coordinates exhibit high-frequency noise and abrupt fluctuations, whereas the filtered trajectories demonstrate smoother and steadier motion. This example illustrates how the IIR Butterworth filter can emphasize tool movement while reducing background jitter.
\section*{Data Availability}
The Cholec80 dataset analyzed in this study is publicly available from the CAMMA laboratory (University Hospital of Strasbourg) under a CC-BY-NC-SA 4.0 license. Access can be requested at https://camma.unistra.fr/datasets/. All additional data generated or analyzed during this study are included in this published article and its supplementary information files.

\section*{Code Availability}
The source code and pretrained models that support the findings of this study are available at: https://github.com/leonlha/KAFR2.

\section*{Acknowledgements}
This work was supported by the National Institutes of Health, National Institute of Biomedical Imaging and Bioengineering, under grant R01EB025247.

\section*{Author contributions statement}
H.P.N.: Conceptualization, Methodology, Coding, Experiments, Validation, Formal analysis, Investigation, Writing - original draft, Writing - review and editing; S.M.K.: Data curation; G.S.: Writing - review and editing, Supervision. All authors have reviewed the manuscript.

\section*{Competing interests}
G.S. is a member of the Editorial Board of npj Digital Surgery. G.S. was not involved in the journal's review of, or decisions related to, this manuscript. The other authors declare no competing financial or non-financial interests.

\bibliography{kafr2.bib}

\end{document}